%% file: root.tex
\documentclass[letterpaper, 10 pt, conference]{ieeeconf}  % Comment this line out if you need a4paper

\IEEEoverridecommandlockouts                              % This command is only needed if 
\usepackage{graphics} % for pdf, bitmapped graphics files
\usepackage{epsfig} % for postscript graphics files
\usepackage{mathptmx} % assumes new font selection scheme installed
\usepackage{times} % assumes new font selection scheme installed
\usepackage{amsmath} % assumes amsmath package installed
\usepackage{amssymb}  % assumes amsmath package installed
\usepackage{diagbox}
\usepackage{bm}
\usepackage{algorithmic}
\usepackage[ruled,vlined]{algorithm2e}
\usepackage{url}

\title{\LARGE \bf
A Novel Approach to Model the Kinematics of Human Fingers Based on an Elliptic Multi-Joint Configuration 
}

\author{Zeyu Wu, Luiza Labazanova, Peng Zhou, and David Navarro-Alarcon% <-this % stops a space
\thanks{This work is supported in part by: The Research Grants Council of Hong Kong (grants 14203917, F-PolyU503/18), the Jiangsu Industrial  Technology Research Institute Collaborative Research Program Scheme (grant ZG9V), the 2019/20 Belt and Road Scholarship (Research Postgraduate), and PolyU (grants G-YBYT, 4-ZZHJ). \textit{Corresponding author: Luiza Labazanova.}}% <-this % stops a space
\thanks{All authors are with the Department of Mechanical Engineering, The Hong Kong Polytechnic University (PolyU), Kowloon, Hong Kong. (e-mail luiza.labazanova@connect.polyu.hk) }%%
}

\begin{document}

\maketitle
\thispagestyle{empty}
\pagestyle{empty}

%%%%%%%%%%%%%%%%%%%%%%%%%%%%%%%%%%%%%%%%%%%%%%%%%%%%%%%%%%%%%%%%%%%%%%%%%%%%%%%%
\begin{abstract}
In this paper, we present a novel kinematic model of the human phalanges based on the elliptical motion of their joints. The presence of the soft elastic tissues and the general anatomical structure of the hand joints highly affect the relative movement of the bones. Commonly used assumption of circular trajectories simplifies the designing process but leads to divergence with the actual hand behavior. The advantages of the proposed model are demonstrated through the comparison with the conventional revolute joint model. Conducted simulations and experiments validate designed forward and inverse kinematic algorithms. Obtained results show a high performance of the model in mimicking the human fingertip motion trajectory.

\end{abstract}

%%%%%%%%%%%%%%%%%%%%%%%%%%%%%%%%%%%%%%%%%%%%%%%%%%%%%%%%%%%%%%%%%%%%%%%%%%%%%%%%
\section{INTRODUCTION}
\input{Sections/introduction}

\section{MODELING OF A BIOMIMETIC HYBRID ROBOTIC FINGER}
\subsection{The Kinematic Model of an Elliptic Joint}
\input{Sections/Methods/Non-circular-description}

\subsection{Forward Kinematics of the Finger}
\input{Sections/Methods/F_Kinematics_Multiplejoint}

\subsection{Inverse Kinematics of the Finger}
\input{Sections/Methods/I_Kinematics_Multiplejoint}

%\subsection{Moment of Elliptical Joints}
%\input{Sections/Methods/Moment_Ellipse}

\section{RESULTS}
In this section, we will present experiment results of the forward and inverse kinematics of one finger with three ellipse joints. The simulations were done in Matlab 2019a on a laptop with 32GB RAM and Intel i7-8500H CPU. All code is available on \url{https://github.com/zhifangsixia/Ellipse_Joint_Finger_Simulation}
\subsection{The Forward Kinematics Simulation}
\input{Sections/Simulation/Fk_simulation}

\subsection{The Inverse Kinematics Simulation}
\input{Sections/Simulation/Ik_simulation}

%\section{EXPERIMENT}
%\input{Sections/Methods}

\subsection{Comparison of Fitting Experiment}
\input{Comparsion_experiment}

\subsection{Leap Motion with Index Fingers}
\input{Leapmotion}

%\subsection{Mechanical Advantage: Elliptical vs Circular}

%\subsection{Exp 3}

\section{CONCLUSIONS AND FUTURE WORK}
\input{Sections/Conclusion}

\addtolength{\textheight}{-12cm}   % This command serves to balance the column lengths
                                  % on the last page of the document manually. It shortens
                                  % the textheight of the last page by a suitable amount.
                                  % This command does not take effect until the next page
                                  % so it should come on the page before the last. Make
                                  % sure that you do not shorten the textheight too much.

%%%%%%%%%%%%%%%%%%%%%%%%%%%%%%%%%%%%%%%%%%%%%%%%%%%%%%%%%%%%%%%%%%%%%%%%%%%%%%%%

%%%%%%%%%%%%%%%%%%%%%%%%%%%%%%%%%%%%%%%%%%%%%%%%%%%%%%%%%%%%%%%%%%%%%%%%%%%%%%%%

%%%%%%%%%%%%%%%%%%%%%%%%%%%%%%%%%%%%%%%%%%%%%%%%%%%%%%%%%%%%%%%%%%%%%%%%%%%%%%%%

%%%%%%%%%%%%%%%%%%%%%%%%%%%%%%%%%%%%%%%%%%%%%%%%%%%%%%%%%%%%%%%%%%%%%%%%%%%%%%%%

\bibliographystyle{ieeetr}
\bibliography{Cite.bib}

\end{document}

%% file: Sections/introduction.tex
The human hand is an essential tool that is used to interact with the surroundings. Hand's dexterity and precision make it highly effective in numerous grasping, manipulation and communication tasks that occur in people routines every day. Researchers have been trying to implement human hand abilities in humanoid robots that demonstrated advanced development in locomotion but lacked upper limbs dexterity. There are multiple examples of bionic robotic hands (BRH) aiming to mimic the hand's shape and general structure. However, their performance is far from their counterpart skills capacity. We suppose that deeper investigation and understanding of the arm's anatomy and functionality is required in order to fulfil the gap between the biological and artificial hands. 

Existing BRHs can be divided into two types according to the structure of the joint: Fixed Rotational Joint (FRJ) and Contact Rotational Joint (CRJ). In FRJ, bionic bones are linked by a rigid shaft; thus, only pure rotation occurs between them. Such a joint has one degree of freedom (DoF), which determines its common use in bionic applications \cite{DLR, bizony_2010_robonaut, ACT}. In CRJ, the articulation surfaces of the bones are connected by soft tissues. Therefore, the bones are not strictly constrained relative to each other, enabling slight extension or compression of the joint \cite{Washington, kim_2019_fluid}.

Both FRJ and CRJ employ the joints that define the circular motion of the output bones. However, there are ongoing disputes about the trajectories generated by the joints within the human hand. In early studies, \cite{rotation1, rotation2, rotation3}, the common assumption was that a circular arc is a proper approximation, whereas more recent anatomical investigations revealed that elliptical paths of the joints are significant to be neglected \cite{ellipse1, ellipse2, ellipse3}. Therefore, in this work, we suggest developing a general kinematic model that will describe the rotation and translation of the joints with non-circular trajectories.

There are various studies striving to implement elliptical motion on the other human joints, such as the shoulder \cite{ellipse_shoulder} and knee \cite{ellipse_knee}. These are examples of single joints while human fingers can be considered as kinematic chains and, thus, their modelling is more complicated. One of the attempts to solve this issue is presented in \cite{ellipse_finger}. However, this work presumes that ellipsoids are in contact with each other. On the other hand, the biological finger joints are composed of one bone with a convex (proximal) head and another one with a concave (distal) head, which are separated by synovial fluid.

\begin{figure}[!ht]
\centering
\includegraphics[width=\columnwidth]{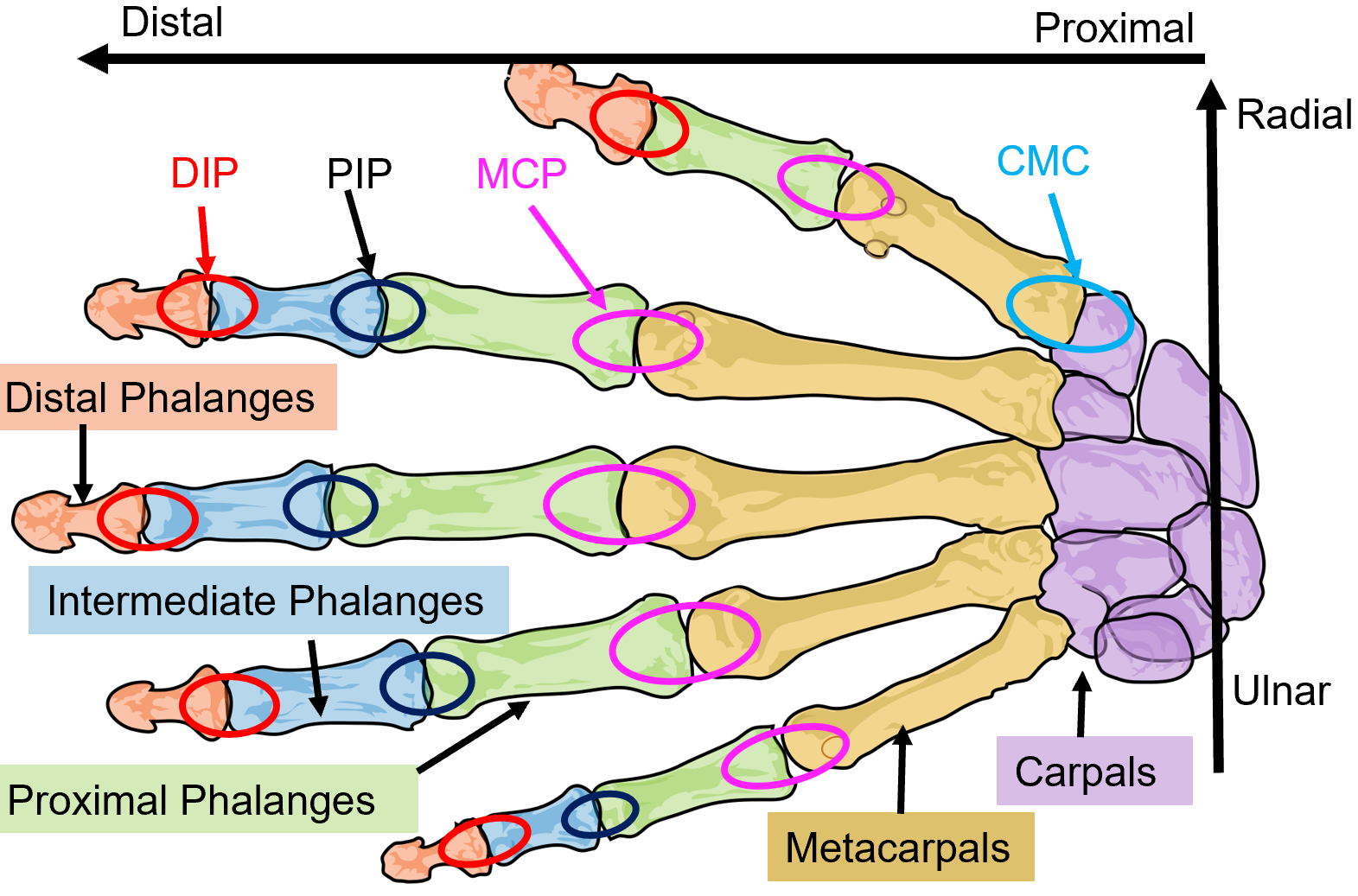}
\caption{The structure of bones of human right hand \cite{structurebones}}
\label{fig: struturebone}
\end{figure}

In this paper, we aim to present a novel kinematic model of the human finger based on the anatomical features of its joints (see Fig. \ref{fig: struturebone}). A multi-joint finger model is designed with the assumption that joints exhibit an elliptic arc motion. The model was further adjusted by curve-fitting with data collected during the movement of the human finger. The main contributions of this work are:

\begin{itemize}
    \item Analytical design of the finger joint that enables elliptical trajectories.
    \item Solution of the forward and inverse kinematics equations that describe the motion of the multi-finger system.
    \item Simulations and experiments that evaluate the performance of the proposed model.
\end{itemize}

%The rest of the manuscript is organised as follows: Sec. II describes the process of kinematics model forming; Sec. III  shows the results and analysis of experiments. Sec. IV makes a conclusion of this paper.

%% file: Sections/Methods/Non-circular-description.tex
This section introduces the designing approach of a kinematic model of the joints with non-circular motion.

\begin{figure}[!ht]
\centering
\includegraphics[scale=0.63]{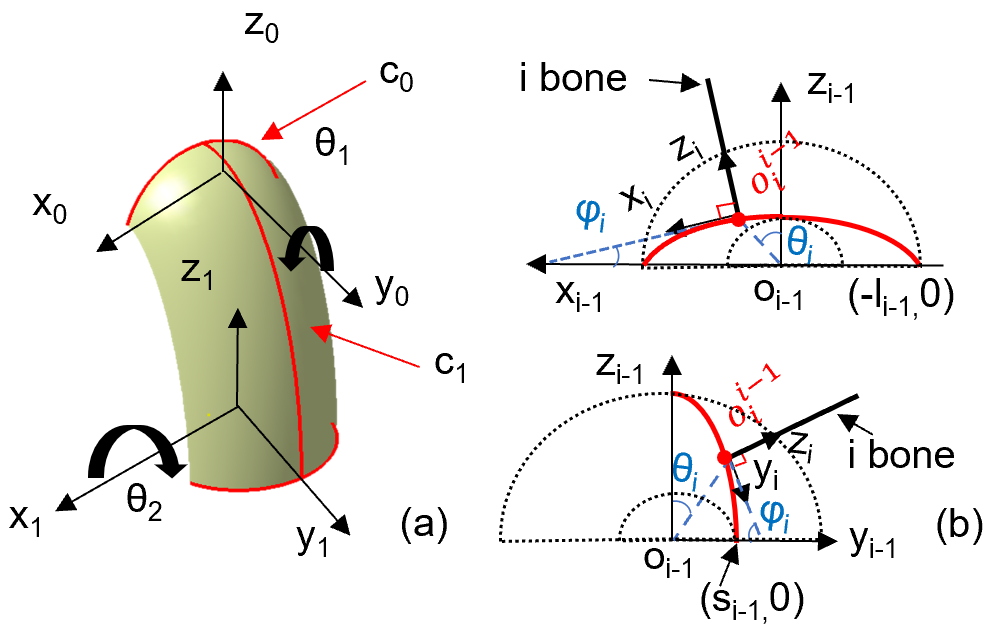}
\caption{The trajectory of elliptical joint: (a) The workspace of MCP joint. (b) the trajectory of joint on xz plane and yz plane.}
\label{fig: singel_ellipse}
\end{figure}

Fig. \ref{fig: singel_ellipse}(a) shows the workspace of a metacarpophalangeal (MCP) joint, which is formed by elliptic trajectories ${{c}_{0}}$ (adduction/abduction) and ${{c}_{1}}$ (flexion/extension). According to the geometric relationships in Fig. \ref{fig: singel_ellipse}(b), the adduction and abduction (Ad/Ab) movements occur on the ${{x}_{i-1}}{{z}_{i-1}}$ plane, and the flexion and extension movements act on the ${{y}_{i-1}}{{z}_{i-1}}$ plane. We define ${{o}_{i}^{i-1}}$ as the origin point of the coordinate frame $i$ under coordinate frame $i-1$. Additionally, ${{o}_{i}^{i-1}}$ represents the translation from coordinate frame $i-1$ to $i$, that is ${{o}_{i}^{i-1}}={{\left[ \begin{matrix}
   {{x}_{i-1}} & {{y}_{i-1}} & {{z}_{i-1}}  \\
\end{matrix} \right]}^{T}}$
$i\in \left[ 1...4 \right]$, which correspond the three joints of the finger. MCP joint can be considered as a combination of the two joints: flexion/extension (F/E) joint and adduction/abduction (Ad/Ab) joint. Thus, the origin point can be determined as follows.

On the ${{x}_{i-1}}{{z}_{i-1}}$ plane:
\begin{equation}
    o_{i}^{i-1}({{\theta }_{i}})={{\left[ \begin{matrix}
   0 & {{z}_{i-1}}\tan ({{\theta }_{i}}) & {{l}_{i-1}}{{s}_{i-1}}\sqrt{\frac{1}{{{\tan }^{2}}{{\theta }_{i}}l_{i-1}^{2}+s_{i-1}^{2}}}  \\
\end{matrix} \text{ }\right]}^{T}}.
\end{equation}

On the ${{y}_{i-1}}{{z}_{i-1}}$ plane:
\begin{equation}
    o_{i}^{i-1}({{\theta }_{i}})\text{=}{{\left[ \begin{matrix}
   {{z}_{i-1}}\tan ({{\theta }_{i}}) & 0 & {{l}_{i-1}}{{s}_{i-1}}\sqrt{\frac{1}{{{\tan }^{2}}{{\theta }_{i}}s_{i-1}^{2}+l_{i-1}^{2}}}  \\
\end{matrix}\text{ } \right]}^{T}},
\end{equation}
where ${{l}_{i-1}}$ and ${{s}_{i-1}}$ are the major and minor axes of the $i$ joint, respectively. We assume that the major axis is on ${{x}_{i-1}}$ and ${{y}_{i-1}}$ according to the anatomy geometry. ${{\theta }_{i}}$ is the rotation angle of the joint, on ${{x}_{i-1}}{{z}_{i-1}}$ plane, ${{\theta }_{i}}\in (0,\pi/2 )$, on ${{y}_{i-1}}{{z}_{i-1}}$ plane, ${{\theta }_{i}}\in (-\pi/2,\pi/2 )$.

According to Fig.\ref{fig: singel_ellipse}(b), ${{\varphi }_{i}}$ is the acute angle formed by the $i$ bone extension line and the horizontal axis, which is the rotation angle of coordinate frame $i-1$ to $i$. It is worth noting that ${{\varphi }_{i}}$ is a function of ${{\theta }_{i}}$, they are not equal except the major axis and minor axis of ellipse are same (at this time, the trajectory is a circle). Their relationship is: 
\begin{align}
  & \text{on }{{x}_{i-1}}{{z}_{i-1}}~\text{plane:  }{{\varphi }_{i}}={{\tan }^{-1}} \left(\frac{s_{i-1}^{2}}{l_{i-1}^{2}}\tan ({{\theta }_{i}})\right) \\ 
 & \text{on }{{y}_{i-1}}{{z}_{i-1}}~\text{plane:  }{{\varphi }_{i}}={{\tan }^{-1}}\left(\frac{l_{i-1}^{2}}{s_{i-1}^{2}}\tan ({{\theta }_{i}})\right) 
\end{align}
The trajectory of the bone has both translation and rotation. We assume a point on coordinate frame $i-1$ is ${{q}_{i-1}}={{\left[ \begin{matrix}
   {{q}_{{{x}_{i-1}}}} & {{q}_{{{y}_{i-1}}}} & {{q}_{{{z}_{i-1}}}}  \\
\end{matrix} \right]}^{T}}$. It can be expressed as:
\begin{align}
  & {{\overset{\_}{\mathop{q}}\,}_{i-1}}=\left[ \begin{matrix}
   {{q}_{i-1}}  \\
   1  \\
\end{matrix} \right]=\left[ \begin{matrix}
   R_{i-1}^{i}({{\varphi }_{i}}) & p_{i-1}^{i}({{\theta }_{i}})  \\
   0 & 1  \\
\end{matrix} \right]\left[ \begin{matrix}
   {{q}_{i}}  \\
   1  \\
\end{matrix} \right]=\overline{T}_{i-1}^{i}{{\overset{\_}{\mathop{q}}\,}_{i}} 
\end{align}

${\overset{\_}{\mathop{q}}_{i-1}\,}$ is the homogeneous coordinate of ${{q}_{i}}$. $\text{   }R_{i-1}^{i}({{\varphi }_{i}})\in SO(3)\text{   }$ is the rotation matrix from $i-1$ to $i$. It's a function of ${{\varphi }_{i}}$. $p_{i-1}^{i}({{\theta }_{i}})\in {{\mathbb{R}}^{3}}$ is the translation from coordinate frame $i-1$ to $i$, which is equal to ${{o}_{i}^{i-1}({{\theta }_{i}})}$. ${T}_{i-1}^{i}$ is the transformation matrix from $i-1$ to $i$.

%% file: Sections/Methods/F_Kinematics_Multiplejoint.tex
\begin{figure}[!ht]
\centering
\includegraphics[scale=0.56]{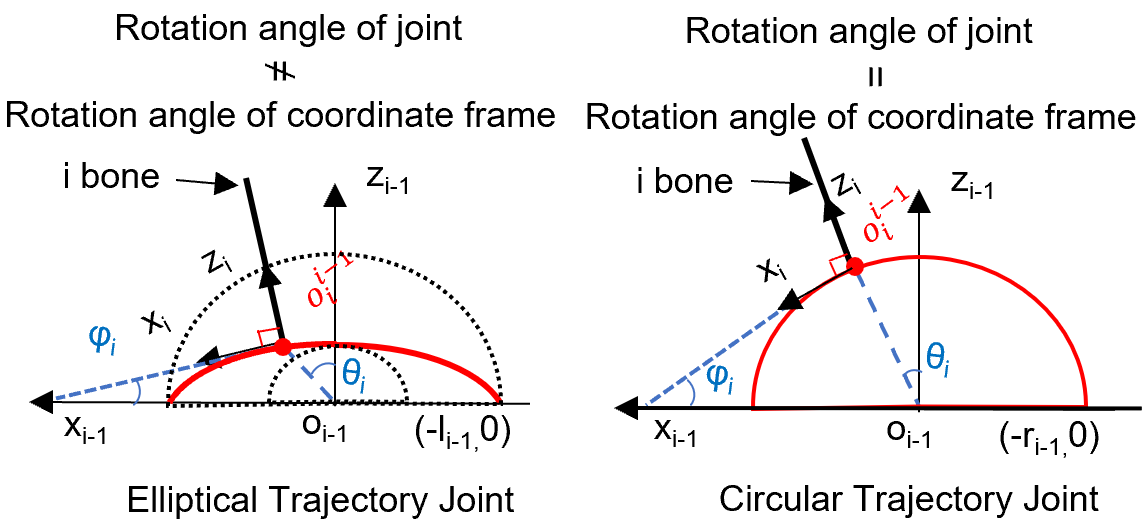}
\caption{The comparison of circular and elliptical joint}
\label{fig: comparsion_circle_ellipse}
\end{figure}

In order to develop the forward kinematics equation of the multi-joints finger, we need to get its transformation matrix. The transformation matrix of a joint with elliptical trajectory (elliptical joint) is different from that of a joint with circular trajectory (circular joint). Generally, for getting the transformation matrix, we need to obtain the rotation angle of the coordinate frame ($\varphi_{i}$) of each joint relative to the previous joint, which we name it as the rotation angle of the coordinate frame. In circular joint, we consider that it equals to the rotation angle of the joint ($\theta_{i}$), which is the angle formed by the intersection of the dashed line and the vertical axis (Fig.\ref{fig: comparsion_circle_ellipse}). Then we input this angle into the joint motor or use it to calculate the displacement of the tendon in tendon-driven system. However, the transformation matrix of elliptical joint is different. It's rotation angle of joint is not equal to rotation angle of coordinate frame. This is because the trajectory of its motion is not only pure rotation, but also having the translation in plane, and the translation is a function of the rotation angle of joint. This means that the translation will change with the rotation of the joint. This is the most significant difference between the elliptical and circular joint. We will introduce how we develop the forward kinematics model of the multi-joints finger with elliptical trajectory joints as follows. 

\begin{figure}[!ht]
\centering
\includegraphics[scale=0.55]{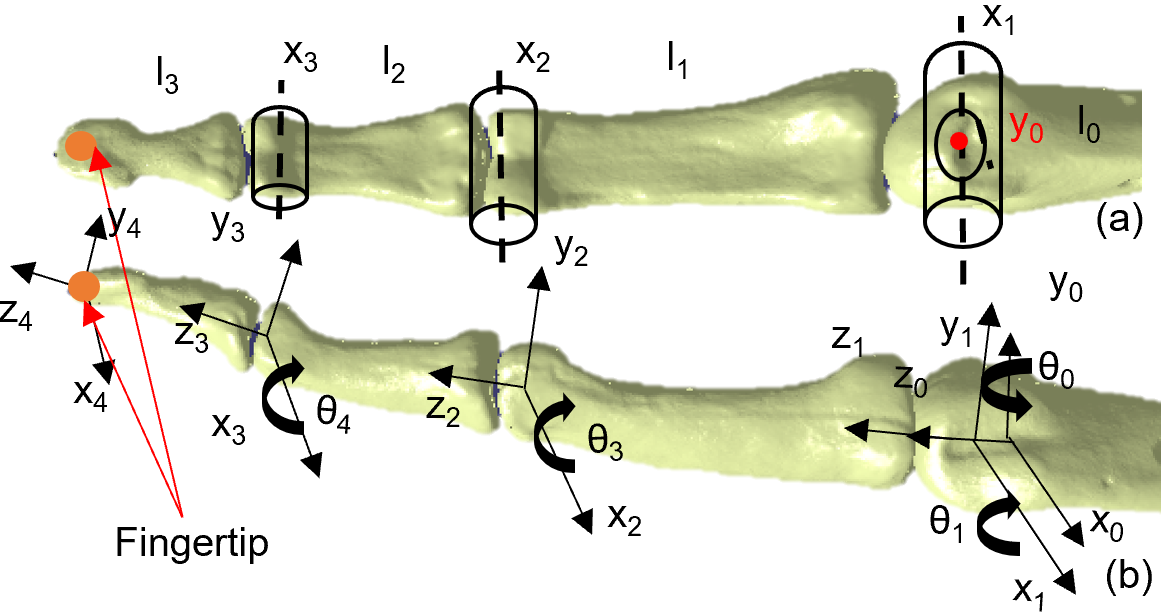}
\caption{The coordinate frame and rotational axes of finger. (a) The rotational axes (b) The coordinate frame from 0-3.}
\label{fig: multi_frame}
\end{figure}

Fig.\ref{fig: multi_frame} (b) shows that the coordinate frames of the finger. There are three condyloid joints with four movements. We choose coordinate frame $0$ represents the Ad/Ab movement, frame $1$ as the F/E of MCP joint, frame $2$ and $3$ as the F/E of PIP and DIP joints. ${l}_{i-1}$ is the length of bone. As equation (5), we can calculate $\overline{T}_{i-1}^{i}$, the first transformation matrix is on ${{x}_{i-1}}{{z}_{i-1}}$ plane, the rest are on on ${{y}_{i-1}}{{z}_{i-1}}$ plane. So we have:
\begin{align}
  & \overline{T}_{0}^{1}={{\left[ \begin{matrix}
   R_{0}^{1}({{\varphi }_{1}}) & o_{0}^{1}({{\theta }_{1}})  \\
   0 & 1  \\
\end{matrix} \right]}_{4\times 4}}\text{ } \\ 
 & \overline{T}_{i-1}^{i}={{\left[ \begin{matrix}
   R_{i-1}^{i}({{\varphi }_{i}}) & o_{i-1}^{i}({{\theta }_{i}})+\left[ \begin{matrix}
   0  \\
   0  \\
   {{l}_{i-1}}  \\
\end{matrix} \right]  \\
   0 & 1  \\
\end{matrix} \right]}_{4\times 4}}\text{(i=2}...\text{4) } 
\end{align}

%${a}_{i-1}$, ${b}_{i-1}$, ${c}_{i-1}$ are the constants on ${x}_{i-1}$, ${y}_{i-1}$, ${z}_{i-1}$ axes to represent the length of long/short axes of ellipse. 
After we have all transformation matrix, we can get the forward kinematics expression:
\begin{equation}
{{\overline{q}}_{0}}=(\prod\limits_{i=1}^{4}{\overline{T}_{i-1}^{i}){{\overline{q}}_{4}}}\text{   i}\in \text{ }\!\![\!\!\text{ 1,4 }\!\!]\!\!\text{ }
\end{equation}
${{\overline{q}}_{0}}$ and ${{\overline{q}}_{4}}$ are the fingertip homogeneous coordinate in coordinate frame $0$ and $4$, ${{\overline{q}}_{4}}=[0,0,0,1]^{T}$.

%% file: Sections/Methods/I_Kinematics_Multiplejoint.tex
\begin{figure}[!ht]
\centering
\includegraphics[scale=0.65]{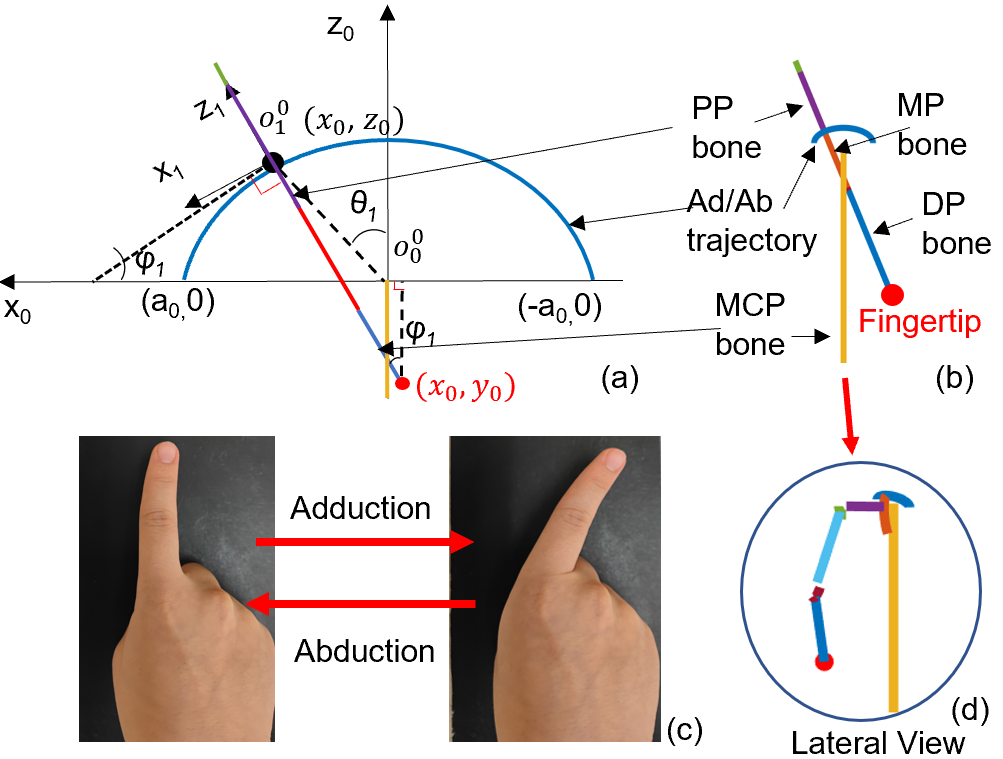}
\caption{The geometry of MCP Ad/Ab trajectory and fingertip. (a) The projection of finger model on ${{x}_{0}}{{z}_{0}}$ plane. (b) The posture of finger model. (c) The Ad/Ab movement of human finger. (d) The lateral side of finger model.} 
\label{fig: theta}
\end{figure}
We have applied three methods of inverse kinematics. The first method refers the work done by Jiewen Lai et al. \cite{lai_2021_verticalizedtip}, firstly input random rotation angle of joints and get the fingertip position by forward kinematics, then save all the data as Point Cloud Library (PCL) with fingertip position and rotation angle of joints. When doing inverse kinematics, we need to search the approximate fingertip position in PCL and return rotation angles. 

The second method is to get the analytical solution of inverse kinematics. We firstly solve the rotation angle of Ad/Ab by using it's physical properties (Fig.\ref{fig: theta}). During the movement of finger (except thumb), Ad/Ab movement occurs on ${{x}_{i-1}}{{z}_{i-1}}$ plane and it's the only movement on this plane. Because we have set it on first coordinate frame, we can directly solve it according to fingertip position ${{\overline{f}}}=[f_{x},f_{y},f_{z}]^{T}$. ${{\theta }_{1}}$ can be expressed as:
\begin{equation}
    F_{1}({{\theta }_{1}})={{\left[ \begin{matrix}
   {{f}_{x}}  \\
   {{f}_{y}}  \\
   {{f}_{z}}  \\
   1  \\
\end{matrix} \right]}^{T}}\left[ \begin{matrix}
   1  \\
   -\tan ({{\varphi }_{1}}({{\theta }_{1}}))  \\
   0  \\
   {{z}_{0}}({{\theta }_{1}})(\tan ({{\varphi }_{1}}({{\theta }_{1}}))-\tan ({{\theta }_{1}}))  \\
\end{matrix} \right]
\end{equation}

${{z}_{0}}({{\theta }_{1}})$ is the ${{Z}_{0}}$ axis coordinate of $o_{1}^{0}$, which is a function of ${{\theta }_{1}}$. $F_{1}({{\theta }_{1}})$ equal to zero and has complex expression which is difficult to solve directly. We use Newton-Raphson method to get the root ($F_{1}({{\theta }_{1}})=0$) of it. The iteration process is:
\begin{equation}
    \Delta {{\theta }_{1}}={{J}^{T}}{{(J{{J}^{T}})}^{-1}}\overset{\to }{\mathop{e}}\,={{J}^{\dagger }}\overset{\to }{\mathop{e}}\
\end{equation}
$\Delta {{\theta }_{1}}$ is the change value of ${{\theta }_{1}}$ for each iteration ($\vartriangle {{\theta }_{1}}=\theta _{1}^{j+1}-\theta _{1}^{j}$), ${j}$ is the increment of iteration. $J$ is the Jacobian matrix, $J=\frac{\partial F({{\theta }_{1}})}{\partial {{\theta }_{1}}}$. ${J}^{\dagger }$ is the Jacobian Pseudo-inverse and ${\overset{\to }{\mathop{e}}}=F({{\theta }_{1}})-0$.

After calculating ${{\theta }_{1}}$, we need to solve the inverse kinematics of F/E of three joints. From equation (9), we can get:
\begin{equation}
{{F}_{2}}(\overline{\theta })=\prod\limits_{i=2}^{4}{T_{i-1}^{i}-{{(\overline{T}_{0}^{1})}^{-1}}{{\overline{q}}_{0}}{{({{\overline{q}}_{4}})}^{-1}}\text{ }i\in [2,4]}
\end{equation}
${{F}_{2}}$ is function of $\overline{\theta }$ and equals to zero. $\overline{\theta }$ is a column vector: ${{\left[ \begin{matrix}
   {{\theta }_{2}} & {{\theta }_{3}} & {{\theta }_{4}}  \\
\end{matrix} \right]}^{T}}$. This three movements are all in sagittal plane which can be regarded as the plane motion, making it have infinite solutions. In order to determine the finger configuration in space, we add a constraint according to anatomical structure. In human finger, the movement of DIP and PIP joint are not independent and have constant ratio when doing flexion without load \cite{ellipse3}. So we assume ${{\theta }_{3}}=\alpha {{\theta }_{4}}$, $\alpha$ is the constant ratio. We can get the analytical solution of inverse kinematics from $(12)$, which contain three equations. However, they are complex implicit equation and difficult to be solved. So we use the numerical method again to solve this equation. The equation can be expressed as:
\begin{equation}
\Delta \overline{\theta }={{\hat{J}}^{T}}{{(\hat{J}{{\hat{J}}^{T}})}^{-1}}\overset{\to }{\mathop{{\hat{e}}}}\,={{\hat{J}}^{\dagger }}\overset{\to }{\mathop{{\hat{e}}}}\,
\end{equation}
$\hat J$ is the Jacobian matrix of ${{F}_{2}}(\overline{\theta })$, which is:
\begin{equation}
    J{{(\overline{\theta })}_{j}}=\text{ }{{\left( \frac{\partial {{F}_{2}}{{(\overline{\theta })}_{i}}}{\partial ({{\overline{\theta }}_{j}})} \right)}_{ij}}\text{ }i,j=[1...3]\text{ }
\end{equation}
${{F}_{2}}{{(\overline{\theta })}_{i}}$ means the $i_{th}$ equation from $(12)$, ${{\overline{\theta }}_{j}}$ equals to ${{\theta }_{j+1}}$. ${\hat{J}}^{\dagger }$ represents the Jacobin Pseudo-inverse and ${\hat{{e}}}\,={{F}_{2}}(\overline{\theta })-0$. 

The third method is for getting the numerical solution of inverse kinematics with multiple constraints. In one finger case, it's similar like the second method. For Ad/Ab movement of MCP joint, we still use geometry method to solve it. For the rest three F/E movements, we use optimization-based approach of numerical inverse kinematics solver as follow:
\begin{equation}
\hat{\theta }=
   \arg \min
   \left(
{{\left\| \overline{f}-Fkine\left( {\hat{\theta }} \right) \right\|}^{2}}+{{\left\| \Delta \hat{\theta } \right\|}^{2}}+r{{\left\| {{\theta }_{4}}-\alpha {{\theta }_{3}} \right\|}^{2}}
\right)
\end{equation}
${\hat{\theta }}\,\in {{\mathbb{R}}^{4}}$ equals to ${{\left[ {{\theta }_{1}}\text{ }{{\theta }_{2}}\text{ }{{\theta }_{3}}\text{ }{{\theta }_{4}} \right]}^{T}}$. $Fkine()\in {{\mathbb{R}}^{3}}$ means the forward kinematics function, which can output the fingertip position according to different rotation angle of joints.

%% file: Sections/Simulation/Fk_simulation.tex
\begin{figure}[!ht]
\centering
\includegraphics[scale=0.42]{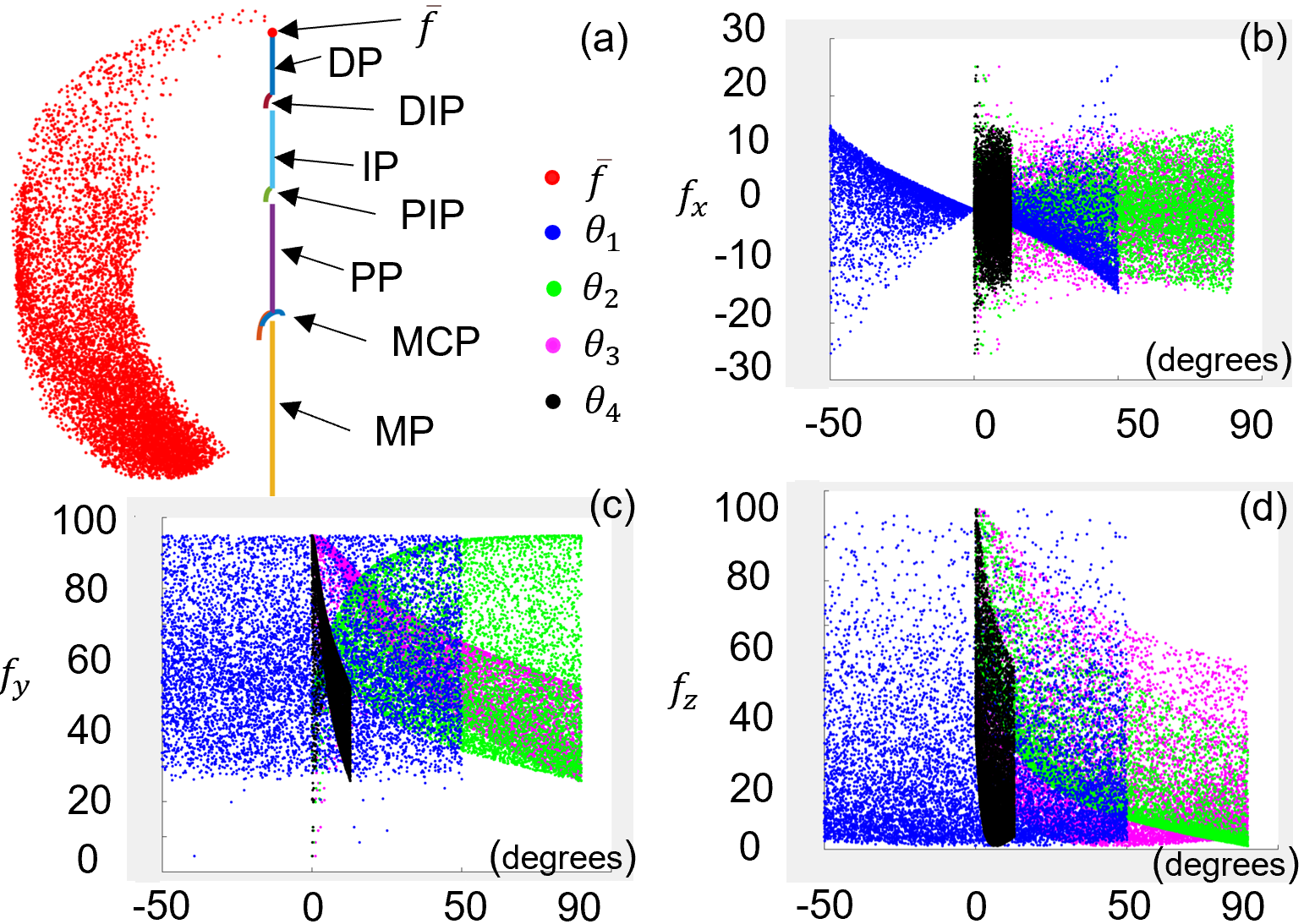}
\caption{The workspace of finger tip: (a) The simulated and the PCL of finger. (b) (c) (d) The relationship of fingertip position and the rotation angle of joints.}
\label{fig: fk1}
\end{figure}

In forward kinematics, we have input the rotation angle of joints. For getting the work space of fingertip, we have used around 8000 sets of $\hat{{{\theta }}}$, the value of input is random (${{\theta }_{1}}\in (-{{50}^{\circ }},{{50}^{\circ }}\text{) }{{\theta }_{2}},{{\theta }_{3}},{{\theta }_{4}}\in ({{0}^{\circ }},{{90}^{\circ }})$, the limits are according to \cite{norkin_2016_measurement}). Using these data we can get the workspace of fingertip (Fig.\ref{fig: fk1}).

\begin{algorithm}
\title{Workspace}
\SetKwInput{KwInput}{Input}                % Set the Input
\SetKwInput{KwOutput}{Output}              % set the Output
\DontPrintSemicolon
  \KwInput{$\hat{\theta }\in {{\mathbb{R}}^{4\times 8000}}\text{ }$ 
  \[(\text{  }\hat{\theta }=[{{\hat{\theta }}_{1}},{{\hat{\theta }}_{2}}...{{\hat{\theta }}_{i}}]\text{   }i\in [1,8000],
  {{\hat{\theta }}_{i}}={{[\theta _{1},\theta _{2},\theta _{3},\theta _{4}]}^{T}})\] 
  $\text{ }para\in {{\hat{R}}^{1\times 12}}$}
  \KwOutput{$fingertip=\left[ \begin{matrix}
   \overline{{{f}_{i}}}  \\
   {{{\hat{\theta }}}_{i}}  \\
\end{matrix} \right]\text{   }fingertip\in {{\mathbb{R}}^{7\times 8000}}$
}
%  \KwData{Testing set $x$}

% Set Function Names
  \SetKwFunction{FMain}{Main}
  \SetKwFunction{FSum}{$\varphi \_T\_Calculate$}
  \SetKwFunction{FSub}{Sub}
 
% Write Function with word ``Function''
  \SetKwProg{Fn}{Function}{:}{}
  \Fn{\FSum{$\theta$, $a$, $b$}}{
         According equation (3), (4) and (5) to calculate\;the transformation matrix.\;
        \KwRet $T$\;
  }
  
% Set Function Names
  \SetKwFunction{FMain}{Main}
  \SetKwFunction{FSum}{$Fkine$}
  \SetKwFunction{FSub}{Sub}
 
% Write Function with word ``Function''
  \SetKwProg{Fn}{Function}{:}{}
  \Fn{\FSum{$\theta$, $para$}}{
             According to $\varphi \_T\_Calculate()$ and equation (9)\;to get the collection of fingertip positions.\;
        \KwRet $\overline{f}$\;
  }

\For{$i=1:8000$}{
$fingertip(1:3,i)={{\hat{\theta }}_{i}}$\;
$fingertip(4:7,i)=Fkine({{\hat{\theta }}_{i}})$\;
}
save $fingertip$
\caption{Workspace of Fingertip}
\end{algorithm}
Algorithm $1$ shows the process of getting the workspace of fingertip. We have defined two functions, the $\varphi \_T\_Calculate()$ is to calculate rotation angle of coordinate frame by rotation angle of joint $\theta$, major axis $a$ and minor axis $b$ of elliptical trajectory. This calculation is according to equation (3), (4) and (5). $T$ is the transformation matrix between coordinate frames. Another function is $Fkine()$, it's for calculating the coordinate of fingertip $\overline{f}$ in coordinate frame $0$. This calculation is according to equation (9). $fingertip$ is a matrix with seven rows and eight thousands columns. The first three rows are the coordinate of fingertip, the rest four rows are the rotation angle of joints.

%% file: Sections/Simulation/Ik_simulation.tex
We have applied two methods to solve the inverse kinematics problem of this finger. The first one is to get the analytical solution of all points on desired trajectory. The second one is searching in the pre-generated PCL. It searches the $\hat{\theta}$ which makes fingertip has minimum Euclidean distance between targets $\overline{f}$ and points in PCL. We have defined two trajectories: the 'heart' trajectory and circle trajectory. We have taken 628 points from 'heart' trajectory and 360 points from circle trajectory, and using two methods to get the inverse kinematics solutions of trajectories. 

In Algorithm 2, we have showed the process of getting analytical solution of inverse kinematics. The input $T$ means the data set of points that we used to form the trajectory. The output $\hat{\theta}$ means the collection of the movement of joint angles. In first function, we have defined a solver for nonlinear equation. The equation should be the form as $F(\theta)=0$. The second function is based on equation (10) and the $NonlinearFunction\_Solver$, we can calculate the movement angle of Ad/Ab by using the $para$ (parameters of finger) and $T$. The third function is to calculate the F/E movement of MCP, PIP and DIP joints based on equation (15). We input the $T$, $para$ and $\hat{\theta_{1}}$ to get the $\hat{\theta}$.

\begin{algorithm}
\title{IK_Analytical}
\SetKwInput{KwInput}{Input}                % Set the Input
\SetKwInput{KwOutput}{Output}              % set the Output
\DontPrintSemicolon
  \KwInput{
  $T\in {{\mathbb{R}}^{3\times n}}\text{ }T=[{{\overline{f}}_{1,}}{{\overline{f}}_{2}}...{{\overline{f}}_{n}}]\text{ };{{\overline{f}}_{n}}={{[f_{x}^{n},f_{y}^{n},f_{z}^{n}]}^{T}}$\;
  $\text{ }\text{ }\text{ }\text{ }\text{ }\text{ }\text{ }\text{ }\text{ }\text{ }para\in {{\mathbb{R}}^{1\times 12}}$
  }
  \KwOutput{
  $\hat{\theta }\in {{\mathbb{R}}^{4\times n}}\text{ }\hat{\theta }={{\left[ {{{\hat{\theta }}}_{1}},{{{\hat{\theta }}}_{2}},{{{\hat{\theta }}}_{3}},{{{\hat{\theta }}}_{4}} \right]}^{T}}$
  $\text{ }{{\hat{\theta }}_{i}}=[\theta _{i}^{1},...\theta _{i}^{n}]\in {{\mathbb{R}}^{1\times n}}\text{ }i\in \left[ 1...4 \right]$
}

% Set Function Names
  \SetKwFunction{FMain}{Main}
  \SetKwFunction{FSum}{$NonlinearFunction\_Solver$}
  \SetKwFunction{FSub}{Sub}
 
% Write Function with word ``Function''
  \SetKwProg{Fn}{Function}{:}{}
  \Fn{\FSum{$F(\theta )$, $\theta$}}{
    Set the error we can bear: $error$; Using equation\;(13) to do iterations until $error$ is enough small.\;
    \KwRet $\theta$\;
  }
% Set Function Names
  \SetKwFunction{FMain}{Main}
  \SetKwFunction{FSum}{$theta1\_calculate$}
  \SetKwFunction{FSub}{Sub}
 
% Write Function with word ``Function''
  \SetKwProg{Fn}{Function}{:}{}
  \Fn{\FSum{$T$, $para$}}{
        Using $NonlinearFunction\_Solver$ and equation\;(10) to calculate the $\hat{\theta }_{1}$\;
        
    \KwRet $\hat{\theta }_{1}$\;
  }
% Set Function Names
  \SetKwFunction{FMain}{Main}
  \SetKwFunction{FSum}{$theta234\_calculate$}
  \SetKwFunction{FSub}{Sub}
 
% Write Function with word ``Function''
  \SetKwProg{Fn}{Function}{:}{}
  \Fn{\FSum{$T$, $para$,$\hat{\theta }_{1}$}}{
        Substitute $\hat{\theta }_{1}$ into $\hat{\theta }$\;
        Get $\overline{f}_{n}$ from $T$\;
        Using $NonlinearFunction\_Solver$ to get the $\hat{\theta }$
    \KwRet $\hat{\theta }$
  }
$\hat{\theta_{1}}=theta1\_calculate(T,para)$\;
$\hat{\theta}=theta234\_calculate(T,para,\hat{\theta_{1}})$\;
save $\hat{\theta}$
\caption{The IK of Analytical Solution}
\end{algorithm}

\begin{algorithm}
\title{IK_PCL}
\SetKwInput{KwInput}{Input}                % Set the Input
\SetKwInput{KwOutput}{Output}              % set the Output
\DontPrintSemicolon
  \KwInput{
  $T\in {{\mathbb{R}}^{3\times n}}\text{ }PCL\in {{\mathbb{R}}^{7\times m}}\text{ }PCL=\left[ {{P}_{1}},{{P}_{2}}...{{P}_{m}} \right]$
  ${{P}_{m}}=\left[ \begin{matrix}
   {{\overline{f}}_{m}}  \\
   {{{\hat{\theta }}}_{m}}  \\
\end{matrix} \right]\text{ }{{\overline{f}}_{m}}=({{x}_{m}},{{y}_{m}},{{z}_{m}})\text{ }{{\hat{\theta }}_{m}}=(\theta _{1}^{m}...\theta _{4}^{m})$

  }
  \KwOutput{
  $\hat{\theta }\in {{\mathbb{R}}^{4\times n}}$
}
\For{$i=1:n$}{
$T\_error=PCL(1:3,:)-T(:,i)\text{ (}T\_error\in {{\mathbb{R}}^{3\times m}})$\;
$error\_norm=sum(T\_error.*T\_error,1)$\;
$(error\_norm\in {{\mathbb{R}}^{1\times m}})$\;
$\left[ \ast{\ },Index(i) \right]=\min (error\_norm)\text{ }Index\in {{\mathbb{R}}^{1\times n}}$
${{\hat{\theta }}_{i}}=PCL\left[ 4:7,Index(i) \right]$\;
}
save $\hat{\theta}$
\caption{The IK of PCL Solution}
\end{algorithm}

\begin{figure}[!ht]
\centering
\includegraphics[scale=0.67]{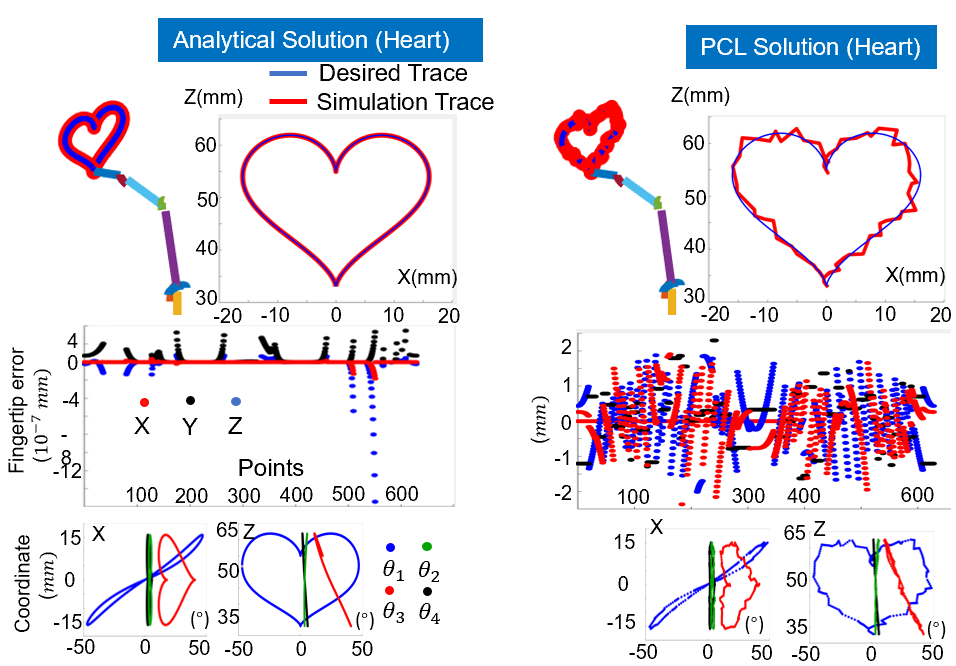}
\caption{Inverse kinematics of analytical solution and PCL solution for 'Heart' trajectory}
\label{fig: fkV1}
\end{figure}
In Algorithm 3, we have expressed that how to find the minimum Euclidean distance between the points in PCL and the target points in $T$. The input are $T$ and PCL. The PCL has $7$ rows and $m$ columns. The first three rows are the coordinate of fingertip, the rest four rows are the corresponding movement angle of joints. $m$ is the amount of points in PCL. For each loop, we calculate the Euclidean distance between the point on trajectory with all points in PCL. $sum(A,1)$ is the function that will sum each column of $A$. $[ \ast{\ },Index]=min(A)$ will return the index (position) of minimum value in $A$. 

\begin{figure}[!ht]
\centering
\includegraphics[scale=0.71]{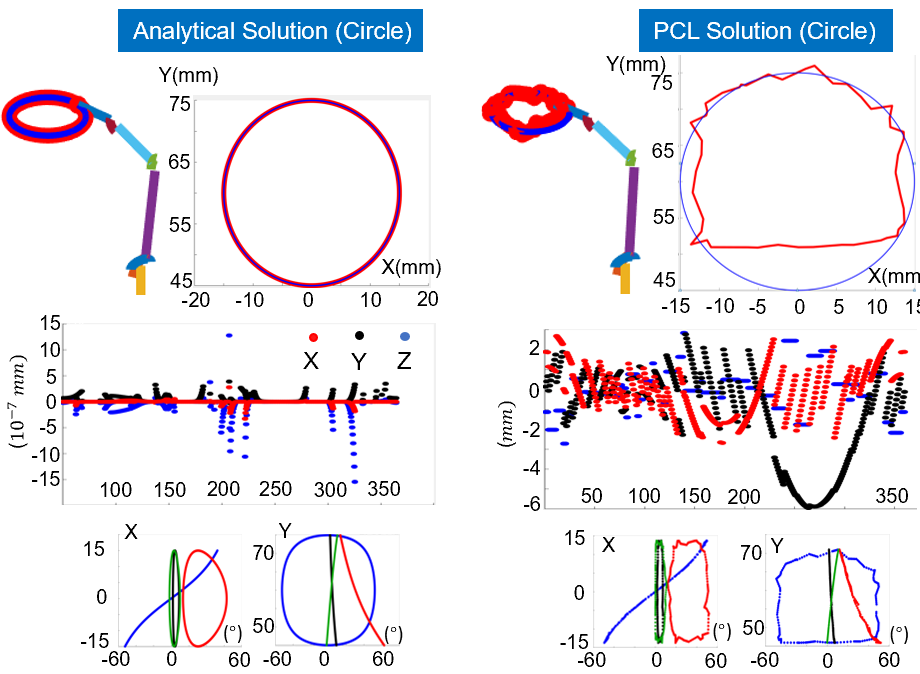}
\caption{Inverse kinematics of analytical solution and PCL solution for circle trajectory}
\label{fig: fkV2}
\end{figure}

Then we can get the simulation result as Fig.\ref{fig: fkV1} and Fig.\ref{fig: fkV2}, the blue curve of 'Heart' and 'Circle' are the desired trajectory that we have generated. According to them, we have done the inverse kinematics in two methods and input the results into $Fkine()$ function to generate the simulation trace. The red curve is the results of simulation. 
In fingertip error charts, the horizontal axis: Points, means the points to form this trace. We have showed the fingertip error between desired and simulation trace. We can find that for analytical solution, it has high accuracy, which the error can be neglected. However, for PCL solution, it has $\pm 2\text{ mm}$ error for 'heart' trace and maximum $6\text{ mm}$ error for circle. At bottom of Fig.\ref{fig: fkV1} and Fig.\ref{fig: fkV2}, we have showed the relationship between the displacement in one axis and the rotation angle of the joints. The heart and circle trace are on the plane which parallels to $XZ\text{ plane}$ and $XY\text{ plane}$ respectively. 

%% file: Comparsion_experiment.tex
\begin{figure}[!ht]
\centering
\includegraphics[width=1\columnwidth]{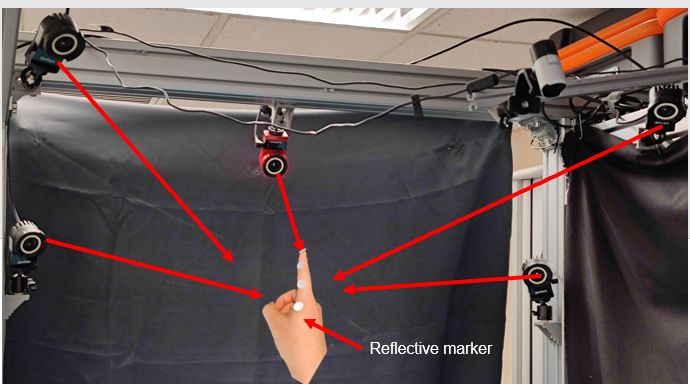}
\caption{The experiment setup}
\label{fig: experiment1_setup}
\end{figure}

\begin{figure}[!ht]
\centering
\includegraphics[width = 1\columnwidth]{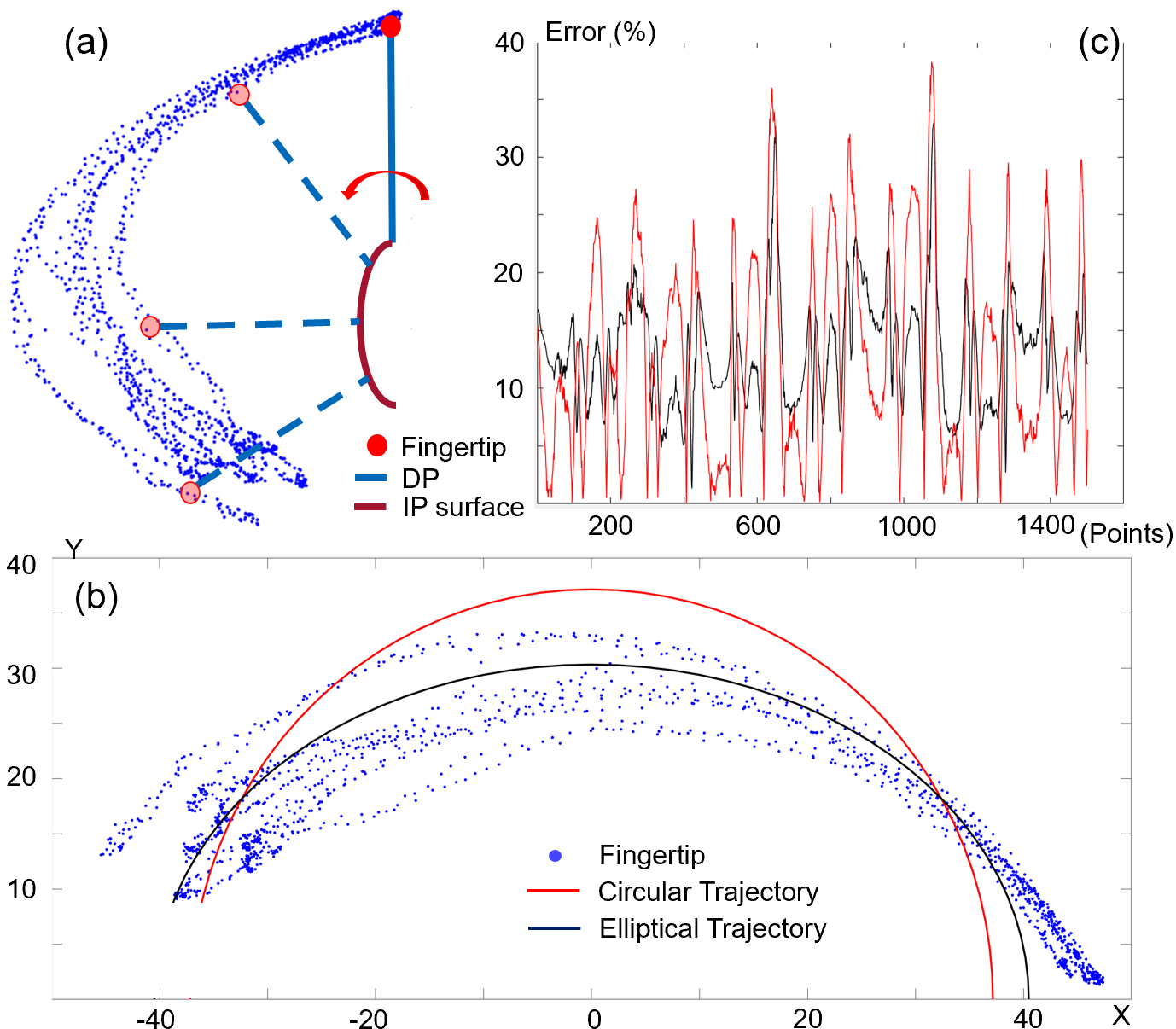}
\caption{The comparison of result fitting by elliptical and circular trajectory: (a) the movement of DIP joint. (b) The error of fitting. (c) The fitting trajectory. }
\label{fig: comparisonTrace}
\end{figure}

In this experiment, we stick reflective markers on index finger, then use OptiTrack cameras to track the markers. We use five OptiTrack cameras to detect 4 markers position (Fig.\ref{fig: experiment1_setup}). One of the markers is fixed on fingertip, and the other three markers are fixed on Intermediate Phalange, Proximal Phalange and Metacarpal of index finger (for locating the finger configuration). The collection of fingertip position is shown in Fig.\ref{fig: comparisonTrace} (a). The blue points are the instantaneous position of the fingertip. It is the movement of the fingertip relative to the Intermediate Phalange, we name it as the movement of DIP joint. We project these points onto a plane to facilitate fitting. In Fig.\ref{fig: comparisonTrace} (b), we have used elliptical and circular trajectories to fit these points. For a single joint, the end (fingertip) trajectory of an elliptical joint is an ellipse, the end trajectory of a circular joint is a circle. Both of them have the same Center of Rotation (CoR), The position of CoR is calculated based on the position of the skin creases of the human index DIP joint. 

Fig.\ref{fig: comparisonTrace} (b) shows the result of fitting. We can see that the fitting performance of the elliptical trajectory is better. Fig.\ref{fig: comparisonTrace} (c) shows the error of fitting. The red curve is the error of circular trajectory and the black curve is the error of elliptical trajectory. We firstly use the actual position of fingertip to get the rotation angle of the joint and calculate the position of the corresponding point on the trajectory, then calculate the distance between the corresponding point and the actual position, and divide by the distance between the corresponding point and the CoR to get the error. The Mean Squared Error (MSE) of elliptical trajectory is 0.0196. The MSE of circular trajectory is 0.0244. The MSE of elliptical trajectory is about $81.7\%$ of circular trajectory, which means the fitting result of the elliptical trajectory has smaller and more stable error.

%% file: Leapmotion.tex
\begin{figure}[!ht]
\centering
\includegraphics[width=1\columnwidth]{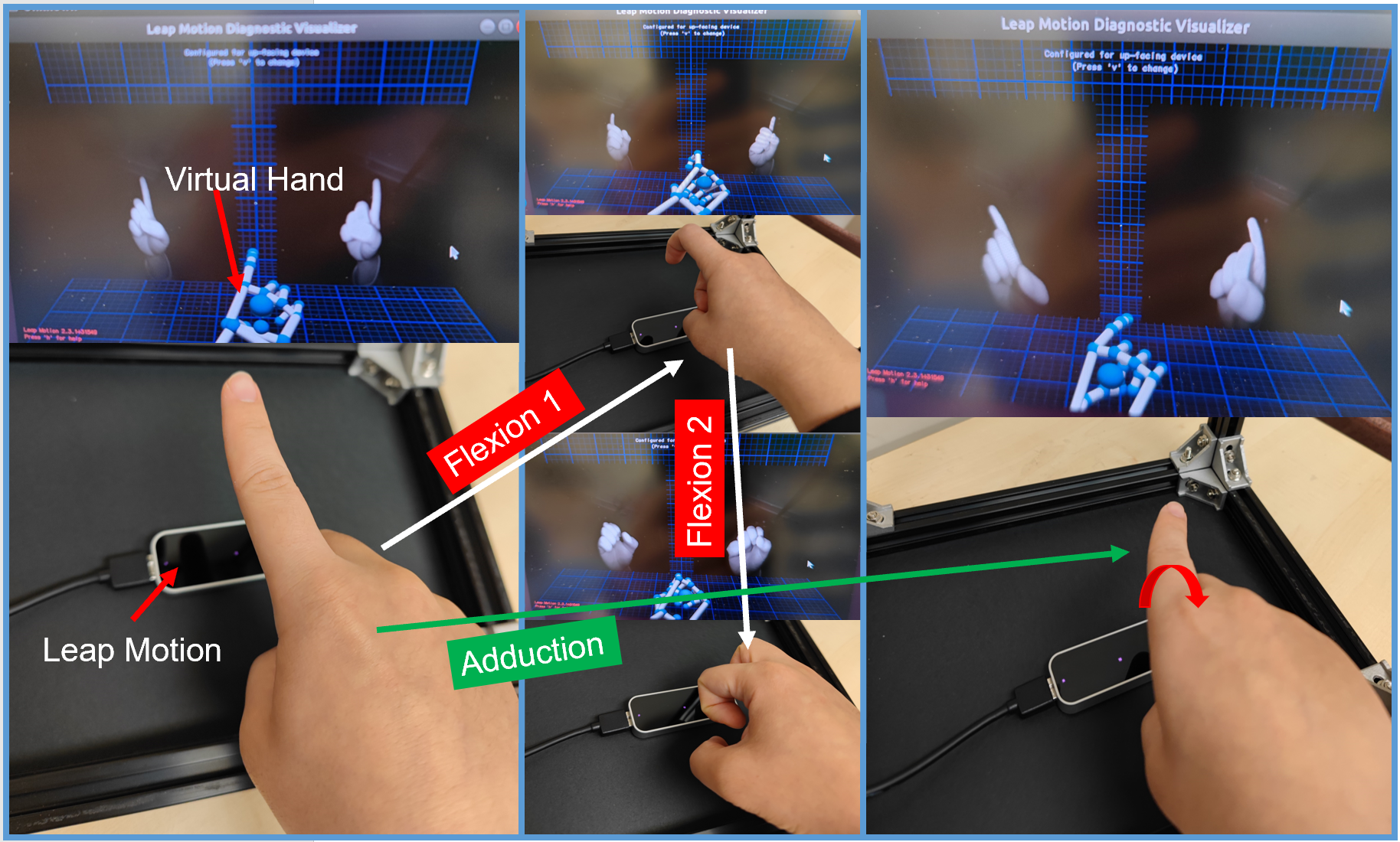}
\caption{The experiment setup: (a) the rest posture. (b) and (c) the process of flexion. (d) the process of adduction. }
\label{fig: experiment_setup}
\end{figure}

After the first experiment, we have conducted the second experiment. In second experiment, we have collected the data of the entire index finger. Because the OptiTrack system can only recognize large markers (compared with finger size) and it is easy to confuse different markers when they are close, we choose Leap Motion to collect the data of finger.

The Leap Motion is able to detect the hand moving and it can return the start and end points coordinate of bones of the fingers. From the data collected we can get the length of bones. We also need to get the length of the major and minor axis of the elliptical trajectory. We let the fingers to do random movement to collect the data, and use nonlinear optimization method to fit the length of the major and minor axis as we have done in previous experiment. 
\begin{figure}[!ht]
\centering
\includegraphics[width=\columnwidth]{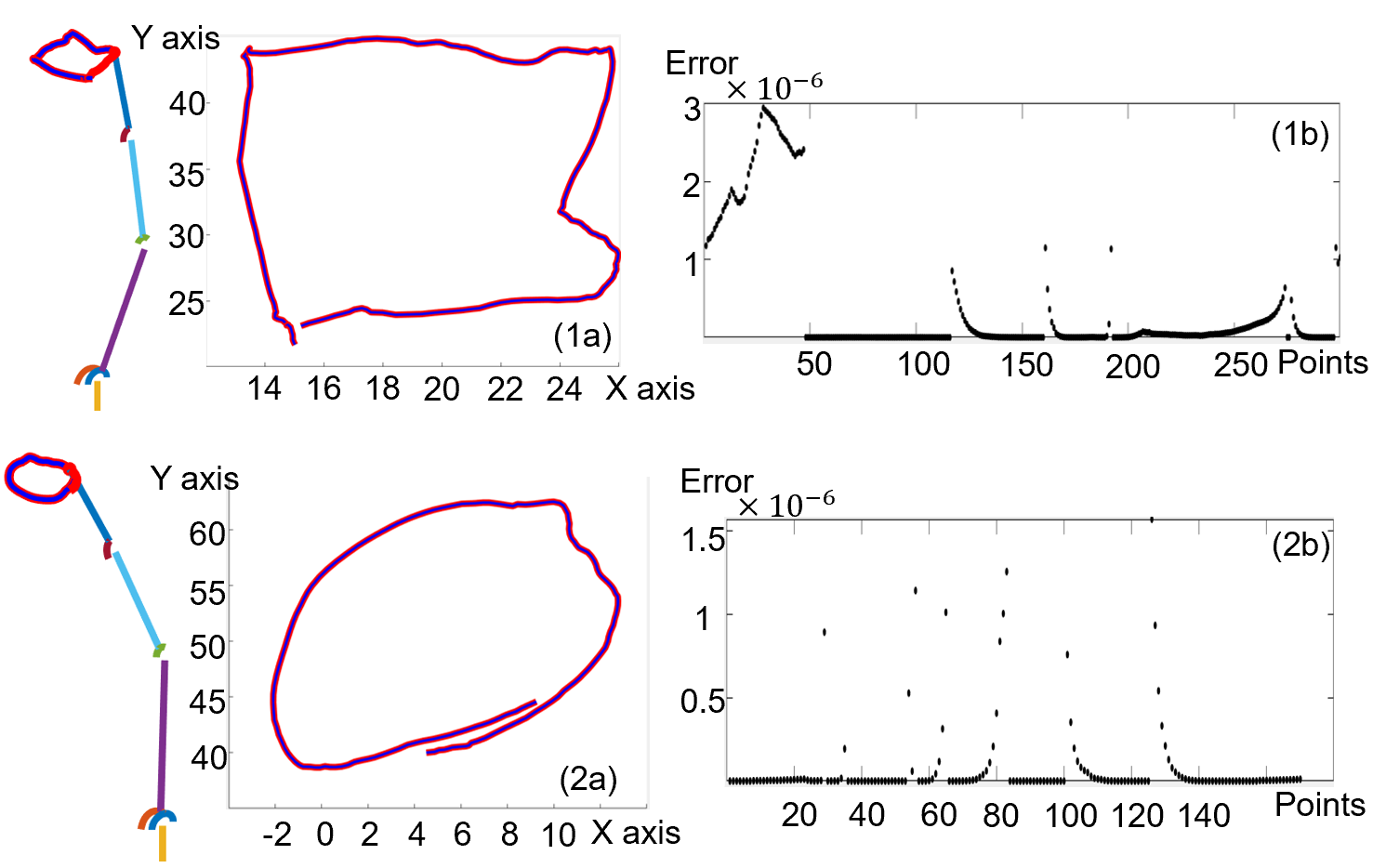}
\caption{The simulation of finger based on real fingertip position}
\label{fig: leapmotionex1}
\end{figure}

In this experiment, we try to make our index finger drawing square and circle. Fig.\ref{fig: leapmotionex1} (1a) and (2a) show the projections of trajectories on plane that we collected from real human index fingertip. We have collected about 300 points for square trajectory and 160 points for circle trajectory. We have done the inverse kinematics according to these two trajectories by using the kinematics model we have fitted in the previous experiment. Then using the result of inverse kinematics to do the forward kinematics to get the simulated fingertip position. The error in Fig.\ref{fig: leapmotionex1} (1b) and (2b) is from the norm of the difference between the real fingertip position and the simulated fingertip position. We can find that for both trajectories, the simulation results have very high accuracy. 

%% file: Sections/Conclusion.tex
In this article, we have proposed a new kinematics model of the multi-joint finger based on elliptical joints, which can better describe the movement of the human finger compared to the conventional circular joints. The error (MSE) of the elliptical joint is $18.3\%$ smaller than that of a circular joint when fitting the human finger joint movement. In our kinematics model, there are four bones with three joints. The Metacarpal is fixed, and the rest bones can move, which are based on joints. Among the joints, DIP and PIP each have one degree of freedom, which is F/E. MCP has two degrees of freedom; one is F/E, the other is Ad/Ab. All joints of the finger are elliptical joints. In our work, we have also used elliptical trajectory to fit the motion of the joint and analyzed the geometric relationship between the rotation angle of the joint and the rotation angle of the coordinate frame in a single joint. Then we have analyzed the difference between our novel elliptical joint and the conventional circular joint. The most obvious difference is the relationship between the rotation angle of the joint and the rotation angle of the coordinate frame. Finally, we have proposed a method for developing a single joint kinematics model.

After that, we have established the forward kinematics model of a multi-joint finger, and then we have developed two methods to get the inverse kinematics solution. The first method is to derive the analytical solution formula of inverse kinematics and then use the numerical method to find the solution of the formula. The second method is to use forward kinematics to generate PCL. When solving inverse kinematics, the algorithm will search for points that are the closest to the target point in PCL and then return the joint angles. 

In the experimental part, we have done four experiments to verify our proposed finger kinematics model. The first one is a forward kinematics simulation experiment. In this experiment, we use forward kinematics to input random joint angles to generate PCL. The second experiment is a simulation experiment of inverse kinematics. We first define two target trajectories: a 'heart' trajectory and the other is a circular trajectory, and then we use two methods we proposed to solve the inverse kinematics of the trajectory. We found that the first method has higher accuracy than the PCL searching method. 

The third experiment is to fit the movement of the DIP joint by using an elliptical and circular joint. The experiment results reveal that the elliptical joint shows high fitting performance, and it has a more minor and more stable error. The last experiment uses a multi-joint finger kinematics model based on elliptical joints to fit the movement of the human index fingertip. We first captured the configuration of random finger movement with Leap Motion and used the nonlinear optimization method to change the major and minor axes of the elliptical joints of our model to fit these data. Then we use a human finger to draw square and circular trajectories and record them by Leap Motion. Then use our fitted kinematics model to solve the inverse kinematics of the trajectories. We found that our simulated fingertip trajectory is similar to the fingertip trajectory collected from the subjects, and the error is minimal, indicating that our kinematics model has high accuracy. In the future, we plan to make a prototype of a robotic finger with elliptical joints to validate our kinematics model.